\documentclass[submission,copyright,creativecommons]{eptcs}
\providecommand{\event}{ICLP-DC 2024}

\usepackage[utf8]{inputenc}
\usepackage{amsmath}
\usepackage{amssymb}
\usepackage{microtype}
\usepackage{url}

\providecommand{\logfont}{\textrm}

\newcommand{\HT}{\ensuremath{\logfont{HT}}}

\newcommand{\HTC}{\ensuremath{\logfont{HT}_{\!c}}}

\newcommand{\LTL}{\ensuremath{\logfont{LTL}}}
\newcommand{\LTLf}{\ensuremath{\LTL_{\!f}}}

\newcommand{\THT}{\ensuremath{\logfont{THT}}}

\newcommand{\TEL}{\ensuremath{\logfont{TEL}}}
\newcommand{\TELf}{\ensuremath{\TEL_{\!f}}}

\newcommand{\LDL}{\ensuremath{\logfont{LDL}}}
\newcommand{\LDLf}{\ensuremath{\LDL_{\!f}}}

\newcommand{\DHT}{\ensuremath{\logfont{DHT}}}

\newcommand{\DEL}{\ensuremath{\logfont{DEL}}}
\newcommand{\DELf}{\ensuremath{\DEL_{\!f}}}

\newcommand{\MEL}{\ensuremath{\logfont{MEL}}}

 \RequirePackage{bm}
\RequirePackage{textcomp}
\RequirePackage{upgreek}

\IfFileExists{outline.tex}{\input{outline}}{}

\newcommand{\next}{\text{\rm \raisebox{-.5pt}{\Large\textopenbullet}}}    

\newcommand{\alwaysF}{\ensuremath{\square}}

\newcommand{\eventuallyF}{\ensuremath{\Diamond}}

\IfFileExists{outline.tex}
        {}
        {}
\IfFileExists{outline.tex}
        {}
        {}
\IfFileExists{outline.tex}
        {}
        {}

\IfFileExists{outline.tex}
        {}
        {}

\mathchardef\mhyphen="2D

 \providecommand{\sysfont}{\textit}

\newcommand{\clingraph}{\sysfont{clingraph}}

\newcommand{\clingcon}{\sysfont{clingcon}}
\newcommand{\clingo}{\sysfont{clingo}}
\newcommand{\clingodl}{\clingoM{dl}}

\newcommand{\telingo}{\sysfont{telingo}}

\newcommand{\clingoM}[1]{\clingo{\small\textnormal{[}\textsc{#1}\textnormal{]}}}

\begin{document}

\newcommand{\AFW}{\ensuremath{\mathrm{AFA}}}

\title{Computational methods for\\Dynamic Answer Set Programming}
\def\titlerunning{Computational methods for Dynamic ASP}
\def\authorrunning{S. Hahn}

\author{Susana Hahn\\
\institute{University of Potsdam, Germany\\
}
\email{hahnmartinlu@uni-potsdam.de}
}

\maketitle

\section{Introduction}\label{sec:introduction}

In our daily lives,
we commonly encounter problems that require reasoning with time.
For instance, planning our day,
determining our route to work,
or scheduling our tasks.
We refer to these problems as 'dynamic' because they involve movement and change over time,
which sometimes includes metric information to express deadlines and durations.
For example, getting to the office within the next hour while ensuring that you have had breakfast beforehand.
In industrial settings, the complexity of these problems increases significantly.
We see this complexity in scenarios such as train scheduling,
production sequencing,
and many other operations.
Therefore, modeling these large-scale problems requires addressing both dynamic aspects
and complex combinatorial optimizations, which is a significant challenge.

Semantic formalisms for expressing dynamic knowledge have been around for many years.
Dynamic logics provide the means to describe ordered events,
making them powerful tools for domains that need to capture actions and changes.
These formalisms are typically approached from a theoretical perspective,
and the systems built around them tend to be single-purpose,
lacking the flexibility to fully model complex problems.
This creates a need for systems that offer comprehensive modeling capabilities for dynamic domains,
efficient solving techniques,
and tools for industrial integration.
Answer Set Programming (ASP) is a prime candidate for solving knowledge-intensive search and optimization problems.
This declarative approach offers a rich modeling language and effective solvers.
However, ASP is primarily suited for static knowledge
and lacks built-in solutions for managing dynamic knowledge.

The overall goal of this research project is to extend ASP into a general-purpose technology for dynamic domains.
The first step is to develop the logical foundations for enhancing ASP's base logic with concepts from dynamic,
temporal,
and metric logic.
Significant progress in this area has already been made by previous efforts of our research group,
providing a solid foundation for further development.
We need to identify fragments of these languages that offer the necessary modeling power for our target dynamic problems
while maintaining properties that allow for formalization and translation using various approaches.
These approaches include using different structures,
such as automata and other transition systems,
and compiling durations into other formalisms,
such as linear constraints.
Implementing these approaches will leverage existing technology in the ASP system \clingo\
and its surrounding tools.
This project will employ advanced programming techniques in ASP to create effective systems for modeling complex dynamic problems.
Additionally,
we aim to add interactive capabilities to these systems to benefit both modelers and end users.
We anticipate that incorporating these features into ASP will enhance users' ability to model dynamic problems
and perform various reasoning tasks.

 \section{Background}\label{sec:background}

\subsection*{Dynamic logics}

One of the most commonly used temporal logics \cite{degola16a} is Linear-time Temporal Logic (LTL) \cite{kamp68a}.
LTL provides modal operators to express temporal properties such as $\next$ (next),
$\alwaysF$ (always),
and $\eventuallyF$ (eventually).
LTL can also be defined in terms of Dynamic Logic (DL) \cite{hatiko00a},
which allows for writing regular expressions over (infinite) traces and mixing declarative with procedural specifications.
These types of specifications are also targeted by action languages such as GOLOG \cite{lereleli97a},
which is based on situation calculus.
Another interesting approach is Metric Temporal Logic (MTL) \cite{koymans90a},
which allows measuring time differences between events.
This measurement is done by assigning a time value to states.
Metric Logic can be used in different applications such as scheduling \cite{luvalibemc16a},
routing \cite{karfra08a},
and more \cite{huozdi20a}.
Originally, these temporal formalisms were investigated for infinite traces.
However, in the past decade, the case of finite traces $\cdot_f$ has gained interest,
as it aligns with a large range of AI applications and constitutes a computationally more feasible variant.
The introduction of $\LTLf$ and Linear Dynamic Logic over finite traces (\LDLf) \cite{giavar13a} served as a stepping stone to define the syntax and classical semantics under this restriction.

There are several tasks addressed by these formalisms and other action theories.
The most commonly known are satisfiability checking,
model checking,
and synthesis \cite{vardi95a, vardi97a, giavar15a, zhpuva19a}.
Furthermore, other more elaborate tasks closer to real-world scenarios include reactive control \cite{batrtu02a},
diagnostics \cite{jiakum04a},
planning \cite{bafrbimc08a, giarub18a, baimci06a},
and verification.

Many of these tasks are solved by translating complex constructs into simpler ones,
for instance, by reducing MTL into LTL \cite{huozdi20a}.
Another very common strategy for addressing these problems is mapping them into automata.
This automata-theoretic approach involves constructing an automaton that accepts exactly the models of a dynamic formula.
This relationship has been extensively researched in areas such as satisfiability checking,
model checking,
planning \cite{bafrbimc08a, giarub18a, baimci06a},
and synthesis \cite{vardi95a, vardi97a, giavar15a, zhpuva19a}.
Non-deterministic finite automata (NFA) \cite{hopull79a} and Deterministic Finite Automata (DFA) have been used for finite traces,
though they are of exponential size relative to the input formula.
To tackle this issue,
\cite{giavar13a} proposed a translation from $\LTLf$ and $\LDLf$ to a more elaborate but succinct automaton:
Alternating Automaton on Finite Words (AFA) \cite{chkost81a, vardi97a, giavar13a}.
These automata, an adaptation of Alternating Büchi Automata to finite traces,
extend NFAs with universal transitions.
This translation, however, led to circular definitions for some dynamic formulas and did not include past operators.
These issues were addressed in \cite{smivar21a},
where the authors introduced Automata Linear Dynamic Logic over finite traces (ALDL$_f$) and presented a translation into even more sophisticated automata:
Two-Way Alternating Finite Automata (2AFA) \cite{laliso84a, kapzak20a}.
In addition to alternation, this type of automaton allows multiple head movements: stationary, left, and right.
More evolved translations from metric logic into automata have also been developed, such as translating MTL into Timed Automata \cite{nivpit10a}.

\subsection*{Answer Set Programming}

Answer Set Programming (ASP) \cite{breitr11a} is a well-established approach to declarative problem-solving
where problems are encoded as logic programs.
The combination of its rich modeling language
and highly effective solving engines makes ASP a very attractive choice.
ASP semantics can be formalized using equilibrium models \cite{pearce06a}
of the logic of here-and-there (HT) \cite{heyting30a}.
This logic has also been extended to here-and-there with constraints (\HTC) \cite{cakaossc16a},
which introduces difference constraints,
a simplified form of linear constraints,
into HT.

The ASP system \clingo\ \cite{gekakasc17a} is known for its high-performing engines.
The system provides various tools for extending the language
and customizing the solving process.
\clingo's theory language capabilities allow for defining custom syntactic expressions.
Additionally, \clingo\ offers two methods for capturing new functionalities \cite{kascwa17a}:
meta-programming,
which uses a reification feature enabling the expression of new functionalities using ASP,
and a sophisticated Python API for manipulating
and customizing the system's internal workflow.
This customization includes techniques such as multi-shot solving,
which allows precise control of the solving process by modularizing the problem.
These features have led to the creation of several hybrid ASP systems.
In particular,
\clingcon~\cite{bakaossc16a}
and \clingodl~\cite{jakaosscscwa17a}
extend the language of \clingo\ with linear constraints using the semantics for \HTC.

\subsection*{Temporal Logic Programming}

In the 1980s, Temporal Logic Programming (TLP) emerged \cite{fukotamo86a, gabbay87b, abaman89a, orgwad92a}.
TLP was revised after the appearance of ASP, resulting in what we know as Temporal ASP.
The idea is to extend the equilibrium models of HT, to deal with dynamic scenarios.
Research began with infinite traces, giving rise to (Linear) Temporal Here-and-There (\THT) \cite{agcadipevi13a}
and (Linear) Dynamic Logic of Here-and-There (\DHT) \cite{bocadisc18a},
along with their non-monotonic counterparts for temporal stable models,
namely Temporal Equilibrium Logic (\TEL) \cite{agcadipevi13a} and Dynamic Equilibrium Logic (\DEL) \cite{cadisc19a}.
The strategy behind these temporal formalisms is to capture time as sequences of HT-interpretations,
resulting in an expressive non-monotonic modal logic.
This approach allowed the definition of Temporal Logic Programs found in \cite{agcadipescscvi20a}.

The temporal operators and semantics of the finite version $\TELf$ were introduced into the ASP system \clingo\,
enriching its modeling power and yielding the first temporal ASP solver \telingo\ \cite{cakamosc19a}.
This system uses the \clingo\ capabilities for theory extensions as well as multi-shot solving in an incremental manner.
Subsequent work incorporated dynamic operators from $\DELf$ \cite{cadisc19a, cadilasc20a}
by unfolding their definitions into $\TELf$ relying on the introduction of auxiliary atoms (in a Tseitin-style \cite{tseitin68a}).
This technique, however, is dependent on a fixed trace length, and the type of translation makes the final logic program hard to interpret.
In \cite{cadiscsc22a}, \TEL\ was further extended with metric temporal operators constrained by intervals over natural numbers,
resulting in Metric Equilibrium Logic (MEL).

 \section{Contributions and future work}\label{sec:discussion}

\subsection*{Automata techniques}

To this moment, I have pursued different translations of dynamic and temporal logic with finite traces into automata,
and implemented them using ASP.
In the current status of the project, I have not yet explored the non-monotonic side of temporal reasoning with automata.
Even though the semantics I have used so far for the temporal formalisms have been monotonic,
I was able to incorporate them in ASP by restricting the dynamic formulas to integrity constraints where their behavior is classical.
With this in mind, at the moment, one can only use these formulas to filter plans via a translation into automata,
which is one of our primary goals.

The first approach, found in \cite{cadihasc21a}, proposes an adaptation of the translation of \LDLf\ to \AFW\ from \cite{giavar13a},
which is incorporated into ASP using meta programming in \clingo.
This implementation is solely based on ASP, relying on the theory extension to define the language for \LDLf\ formulas,
and the reified output of \clingo.
This reification corresponds to a linearized representation of the dynamic formula as facts based on the grammar defined for the theory.
Then, using an ASP encoding, these facts are translated into a declarative representation of the corresponding alternating automaton.
In the full version of this work \cite{cadihasc21b}, we explore other existing tools for computing an automaton from a dynamic formula.
In order to employ them, we developed two different translations from $\LDLf$ to Monadic Second Order Logic (MSO).
For the implementation, we parsed the formula with $\clingo$'s API and called the state-of-the-art system MONA \cite{hejejoklparasa95a} to obtain a DFA,
which is then transformed into facts.
By having a unified declarative representation to capture the different automata ($\AFW$ and DFA),
we were able to craft a single encoding for checking the acceptance of the automata.

Following (soon to be published) work presents a novel translation from \LDLf\ into 2AFA.
Leveraging the transitions without head movement provided by these automata,
we were able to remove the recursive nature of our old translation, thus eliminating the circular issue carried from \cite{giavar13a}.
Furthermore, the left head movement allows us to readily refer to time points in the past.
These new target automata, nonetheless, represented a formalization challenge since there is a lack of literature available in contrast to simpler automata.
This translation was implemented as part of the \texttt{adlingo}\footnote{\url{https://github.com/potassco/adlingo}} system.
Just like the previous work, the implementation was done using meta programming and theory extensions.
Additionally, it integrates the use of \clingraph\ \cite{hasascst24a} to visualize the automata and its runs using an ASP encoding.

\subsection*{Linear constraints to encode durations}

My latest work has focused on constructing the foundations of Metric Logic Programs (MLP).
With these programs, we plan to abstract the basic modalities and forms needed to model metric problems in ASP.
The semantics of these programs are those of \MEL, where, by restricting the syntax, we aim to simplify the computation.
The first approach for this work has been submitted to ICLP24.
As in the automata approach, we are restricting the research to the finite setting for a given (fixed) horizon.
This work defines the basic syntax for a MLP in which all rules are universal, meaning that they must hold in every step.
This contrasts with the approach used for temporal logic programs in \cite{agcadipescscvi20a},
where the rules were separated into initial, dynamic, and final, which facilitates the implementation using incremental solving.
For our approach, the removal of this division simplifies the use of meta-programming techniques to define the translation, as well as the overall semantics.
The use of meta-programming allows us to define the translations in ASP,
making the implementation transparent and modular.
A big advantage of this approach is the clear mapping between the translation and the implementation in ASP.
As a consequence, it simplifies the proves of correctness and completeness of the translation.

In this first exploration, we restricted the rules of metric logic programs to only use the metric next modality.
For instance, with the rule $\next_{[20..40)} \textit{school}\leftarrow \textit{drive}$, we express that \textit{``If I start driving, I must be at the school in the next step, which should happen in 20 to 40 minutes"}.
We suspect that the core of our target dynamic problems can be modeled with this restricted fragment.
In a nutshell, the first part of this translation represents the state changes and follows the same semantics as in $\TELf$.
The second part accounts for the timed aspect of metric logic.
For this part, we explore two approaches: one where the translation is done to \HT, and a second one where the target logic is \HTC.
As a result, we were able to see what we expected: a succinct and performant translation of time into linear constraints.
We also observed that our restricted language did allow us to model the transitions and time restrictions, but was not enough to represent the goal conditions of the problems.
These conditions usually require more complex metric operators to talk about states that are further away in time, for instance, \textit{``At some point in the next 2 hours, I will be back home"}.

\subsection{Future work}

The quest to conceptualize metric logic programming is far from over.
In view of the results from the last project, we have started to craft a translation that handles more metric operators.
The translation is planned to follow a Tseitin-style translation like the one in \cite{agcadipescscvi20a}.
We want to further investigate \HTC\ for encoding time,
and examine the integration of non-monotonic reasoning and optimization in the timed aspect of MLP.
Additionally, we plan to investigate how far ASP can address reactive-dynamic tasks where the user and environment
play a role by interacting with the system.

\bibliographystyle{./include/latex-class-eptcs/eptcs}

\begin{thebibliography}{10}
\providecommand{\bibitemdeclare}[2]{}
\providecommand{\surnamestart}{}
\providecommand{\surnameend}{}
\providecommand{\urlprefix}{Available at }
\providecommand{\url}[1]{\texttt{#1}}
\providecommand{\href}[2]{\texttt{#2}}
\providecommand{\urlalt}[2]{\href{#1}{#2}}
\providecommand{\doi}[1]{doi:\urlalt{http://dx.doi.org/#1}{#1}}
\providecommand{\bibinfo}[2]{#2}

\bibitemdeclare{article}{abaman89a}
\bibitem{abaman89a}
\bibinfo{author}{M.~\surnamestart Abadi\surnameend} \&
  \bibinfo{author}{Z.~\surnamestart Manna\surnameend} (\bibinfo{year}{1989}):
  \emph{\bibinfo{title}{Temporal Logic Programming}}.
\newblock {\sl \bibinfo{journal}{Journal of Symbolic Computation}}
  \bibinfo{volume}{8}, pp. \bibinfo{pages}{277--295},
  \doi{10.1016/S0747-7171(89)80070-7}.

\bibitemdeclare{article}{agcadipescscvi20a}
\bibitem{agcadipescscvi20a}
\bibinfo{author}{F.~\surnamestart Aguado\surnameend},
  \bibinfo{author}{P.~\surnamestart Cabalar\surnameend},
  \bibinfo{author}{M.~\surnamestart Di{\'{e}}guez\surnameend},
  \bibinfo{author}{G.~\surnamestart P{\'{e}}rez\surnameend},
  \bibinfo{author}{T.~\surnamestart Schaub\surnameend},
  \bibinfo{author}{A.~\surnamestart Schuhmann\surnameend} \&
  \bibinfo{author}{C.~\surnamestart Vidal\surnameend} (\bibinfo{year}{2023}):
  \emph{\bibinfo{title}{Linear-Time Temporal Answer Set Programming}}.
\newblock {\sl \bibinfo{journal}{Theory and Practice of Logic Programming}}
  \bibinfo{volume}{23}(\bibinfo{number}{1}), pp. \bibinfo{pages}{2--56},
  \doi{10.1017/S1471068421000557}.

\bibitemdeclare{article}{agcadipevi13a}
\bibitem{agcadipevi13a}
\bibinfo{author}{F.~\surnamestart Aguado\surnameend},
  \bibinfo{author}{P.~\surnamestart Cabalar\surnameend},
  \bibinfo{author}{M.~\surnamestart Di{\'e}guez\surnameend},
  \bibinfo{author}{G.~\surnamestart P{\'e}rez\surnameend} \&
  \bibinfo{author}{C.~\surnamestart Vidal\surnameend} (\bibinfo{year}{2013}):
  \emph{\bibinfo{title}{Temporal equilibrium logic: a survey}}.
\newblock {\sl \bibinfo{journal}{Journal of Applied Non-Classical Logics}}
  \bibinfo{volume}{23}(\bibinfo{number}{1-2}), pp. \bibinfo{pages}{2--24},
  \doi{10.1080/11663081.2013.798985}.

\bibitemdeclare{inproceedings}{bafrbimc08a}
\bibitem{bafrbimc08a}
\bibinfo{author}{J.~\surnamestart Baier\surnameend},
  \bibinfo{author}{C.~\surnamestart Fritz\surnameend},
  \bibinfo{author}{M.~\surnamestart Bienvenu\surnameend} \&
  \bibinfo{author}{S.~\surnamestart McIlraith\surnameend}
  (\bibinfo{year}{2008}): \emph{\bibinfo{title}{Beyond Classical Planning:
  Procedural Control Knowledge and Preferences in State-of-the-Art Planners}}.
\newblock In \bibinfo{editor}{D.~\surnamestart Fox\surnameend} \&
  \bibinfo{editor}{C.~\surnamestart Gomes\surnameend}, editors: {\sl
  \bibinfo{booktitle}{Proceedings of the Twenty-third National Conference on
  Artificial Intelligence (AAAI'08)}}, \bibinfo{publisher}{{AAAI} Press}, pp.
  \bibinfo{pages}{1509--1512}.
\newblock \urlprefix\url{https://auld.aaai.org/Library/AAAI/2008/aaai08-251.php}.

\bibitemdeclare{inproceedings}{baimci06a}
\bibitem{baimci06a}
\bibinfo{author}{J.~\surnamestart Baier\surnameend} \&
  \bibinfo{author}{S.~\surnamestart McIlraith\surnameend}
  (\bibinfo{year}{2006}): \emph{\bibinfo{title}{Planning with First-Order
  Temporally Extended Goals using Heuristic Search}}.
\newblock In \bibinfo{editor}{Y.~\surnamestart Gil\surnameend} \&
  \bibinfo{editor}{R.~\surnamestart Mooney\surnameend}, editors: {\sl
  \bibinfo{booktitle}{Proceedings of the Twenty-first National Conference on
  Artificial Intelligence (AAAI'06)}}, \bibinfo{publisher}{{AAAI} Press}, pp.
  \bibinfo{pages}{788--795}.
\newblock \urlprefix\url{https://www.aaai.org/Library/AAAI/2006/aaai06-125.php}.

\bibitemdeclare{proceedings}{lpnmr19}
\bibitem{lpnmr19}
\bibinfo{editor}{M.~\surnamestart Balduccini\surnameend},
  \bibinfo{editor}{Y.~\surnamestart Lierler\surnameend} \&
  \bibinfo{editor}{S.~\surnamestart Woltran\surnameend}, editors
  (\bibinfo{year}{2019}): \emph{\bibinfo{title}{Proceedings of the Fifteenth
  International Conference on Logic Programming and Nonmonotonic Reasoning
  (LPNMR'19)}}. {\sl \bibinfo{series}{Lecture Notes in Artificial
  Intelligence}} \bibinfo{volume}{11481}, \bibinfo{publisher}{Springer-Verlag},
  \doi{10.1007/978-3-030-20528-7}.

\bibitemdeclare{article}{bakaossc16a}
\bibitem{bakaossc16a}
\bibinfo{author}{M.~\surnamestart Banbara\surnameend},
  \bibinfo{author}{B.~\surnamestart Kaufmann\surnameend},
  \bibinfo{author}{M.~\surnamestart Ostrowski\surnameend} \&
  \bibinfo{author}{T.~\surnamestart Schaub\surnameend} (\bibinfo{year}{2017}):
  \emph{\bibinfo{title}{Clingcon: The Next Generation}}.
\newblock {\sl \bibinfo{journal}{Theory and Practice of Logic Programming}}
  \bibinfo{volume}{17}(\bibinfo{number}{4}), pp. \bibinfo{pages}{408--461},
  \doi{10.1017/S1471068417000138}.

\bibitemdeclare{inproceedings}{batrtu02a}
\bibitem{batrtu02a}
\bibinfo{author}{C.~\surnamestart Baral\surnameend}, \bibinfo{author}{S.~Tran
  \surnamestart Cao\surnameend} \& \bibinfo{author}{L.~\surnamestart
  Tuan\surnameend} (\bibinfo{year}{2002}): \emph{\bibinfo{title}{A transition
  function based characterization of actions with delayed and continuous
  effects}}.
\newblock In: {\sl \bibinfo{booktitle}{KR}}, \bibinfo{organization}{Citeseer},
  pp. \bibinfo{pages}{291--302}.
\newblock \urlprefix\url{https://citeseerx.ist.psu.edu/document?repid=rep1&type=pdf&doi=d4ae7ddfcdc012e519b473d03dd3c7caffaa09e1}.

\bibitemdeclare{inproceedings}{bocadisc18a}
\bibitem{bocadisc18a}
\bibinfo{author}{A.~\surnamestart Bosser\surnameend},
  \bibinfo{author}{P.~\surnamestart Cabalar\surnameend},
  \bibinfo{author}{M.~\surnamestart Di\'eguez\surnameend} \&
  \bibinfo{author}{T.~\surnamestart Schaub\surnameend} (\bibinfo{year}{2018}):
  \emph{\bibinfo{title}{Introducing Temporal Stable Models for Linear Dynamic
  Logic}}.
\newblock In \bibinfo{editor}{M.~\surnamestart Thielscher\surnameend},
  \bibinfo{editor}{F.~\surnamestart Toni\surnameend} \&
  \bibinfo{editor}{F.~\surnamestart Wolter\surnameend}, editors: {\sl
  \bibinfo{booktitle}{Proceedings of the Sixteenth International Conference on
  Principles of Knowledge Representation and Reasoning (KR'18)}},
  \bibinfo{publisher}{{AAAI} Press}, pp. \bibinfo{pages}{12--21}.
\newblock \urlprefix\url{https://aaai.org/ocs/index.php/KR/KR18/paper/view/18047}.

\bibitemdeclare{article}{breitr11a}
\bibitem{breitr11a}
\bibinfo{author}{G.~\surnamestart Brewka\surnameend},
  \bibinfo{author}{T.~\surnamestart Eiter\surnameend} \&
  \bibinfo{author}{M.~\surnamestart Truszczy{\'n}ski\surnameend}
  (\bibinfo{year}{2011}): \emph{\bibinfo{title}{Answer set programming at a
  glance}}.
\newblock {\sl \bibinfo{journal}{Communications of the {ACM}}}
  \bibinfo{volume}{54}(\bibinfo{number}{12}), pp. \bibinfo{pages}{92--103},
  \doi{10.1145/2043174.2043195}.

\bibitemdeclare{inproceedings}{cadihasc21a}
\bibitem{cadihasc21a}
\bibinfo{author}{P.~\surnamestart Cabalar\surnameend},
  \bibinfo{author}{M.~\surnamestart Di{\'{e}}guez\surnameend},
  \bibinfo{author}{S.~\surnamestart Hahn\surnameend} \&
  \bibinfo{author}{T.~\surnamestart Schaub\surnameend} (\bibinfo{year}{2021}):
  \emph{\bibinfo{title}{Automata for Dynamic Answer Set Solving: Preliminary
  Report}}.
\newblock In: {\sl \bibinfo{booktitle}{Proceedings of the Fourteenth Workshop
  on Answer Set Programming and Other Computing Paradigms (ASPOCP'21)}}.
\newblock \urlprefix\url{https://ceur-ws.org/Vol-2970/aspocpinvited1.pdf}.

\bibitemdeclare{article}{cadihasc21b}
\bibitem{cadihasc21b}
\bibinfo{author}{P.~\surnamestart Cabalar\surnameend},
  \bibinfo{author}{M.~\surnamestart Dieguez\surnameend},
  \bibinfo{author}{S.~\surnamestart Hahn\surnameend} \&
  \bibinfo{author}{T.~\surnamestart Schaub\surnameend} (\bibinfo{year}{2021}):
  \emph{\bibinfo{title}{Automata for dynamic answer set solving: Preliminary
  report}}.
\newblock {\sl \bibinfo{journal}{CoRR}} \bibinfo{volume}{abs/2109.01782}, \doi{10.48550/arXiv.2109.01782
}.

\bibitemdeclare{inproceedings}{cadilasc20a}
\bibitem{cadilasc20a}
\bibinfo{author}{P.~\surnamestart Cabalar\surnameend},
  \bibinfo{author}{M.~\surnamestart Di\'eguez\surnameend},
  \bibinfo{author}{F.~\surnamestart Laferriere\surnameend} \&
  \bibinfo{author}{T.~\surnamestart Schaub\surnameend} (\bibinfo{year}{2020}):
  \emph{\bibinfo{title}{Implementing Dynamic Answer Set Programming over finite
  traces}}.
\newblock In \bibinfo{editor}{G.~\surnamestart {De Giacomo}\surnameend},
  \bibinfo{editor}{A.~\surnamestart Catal{\'a}\surnameend},
  \bibinfo{editor}{B.~\surnamestart Dilkina\surnameend},
  \bibinfo{editor}{M.~\surnamestart Milano\surnameend},
  \bibinfo{editor}{S.~\surnamestart Barro\surnameend},
  \bibinfo{editor}{A.~\surnamestart Bugar{\'{\i}}n\surnameend} \&
  \bibinfo{editor}{J.~\surnamestart Lang\surnameend}, editors: {\sl
  \bibinfo{booktitle}{Proceedings of the Twenty-fourth European Conference on
  Artificial Intelligence (ECAI'20)}}, \bibinfo{publisher}{{IOS} Press}, pp.
  \bibinfo{pages}{656--663}, \doi{10.3233/FAIA200151}.

\bibitemdeclare{inproceedings}{cadisc19a}
\bibitem{cadisc19a}
\bibinfo{author}{P.~\surnamestart Cabalar\surnameend},
  \bibinfo{author}{M.~\surnamestart Di{\'e}guez\surnameend} \&
  \bibinfo{author}{T.~\surnamestart Schaub\surnameend} (\bibinfo{year}{2019}):
  \emph{\bibinfo{title}{Towards Dynamic Answer Set Programming over finite
  traces}}.
\newblock In \bibinfo{editor}{Balduccini} et~al.  \cite{lpnmr19}, pp.
  \bibinfo{pages}{148--162}, \doi{10.1007/978-3-030-20528-7\_12}.

\bibitemdeclare{inproceedings}{cadiscsc22a}
\bibitem{cadiscsc22a}
\bibinfo{author}{P.~\surnamestart Cabalar\surnameend},
  \bibinfo{author}{M.~\surnamestart Di{\'{e}}guez\surnameend},
  \bibinfo{author}{T.~\surnamestart Schaub\surnameend} \&
  \bibinfo{author}{A.~\surnamestart Schuhmann\surnameend}
  (\bibinfo{year}{2022}): \emph{\bibinfo{title}{Metric Temporal Answer Set
  Programming over Timed Traces}}.
\newblock In \bibinfo{editor}{G.~\surnamestart Gottlob\surnameend},
  \bibinfo{editor}{D.~\surnamestart Inclezan\surnameend} \&
  \bibinfo{editor}{M.~\surnamestart Maratea\surnameend}, editors: {\sl
  \bibinfo{booktitle}{Proceedings of the Sixteenth International Conference on
  Logic Programming and Nonmonotonic Reasoning (LPNMR'22)}}, {\sl
  \bibinfo{series}{Lecture Notes in Artificial Intelligence}}
  \bibinfo{volume}{13416}, \bibinfo{publisher}{Springer-Verlag}, pp.
  \bibinfo{pages}{117--130}, \doi{10.1007/978-3-031-15707-3\_10}.

\bibitemdeclare{inproceedings}{cakamosc19a}
\bibitem{cakamosc19a}
\bibinfo{author}{P.~\surnamestart Cabalar\surnameend},
  \bibinfo{author}{R.~\surnamestart Kaminski\surnameend},
  \bibinfo{author}{P.~\surnamestart Morkisch\surnameend} \&
  \bibinfo{author}{T.~\surnamestart Schaub\surnameend} (\bibinfo{year}{2019}):
  \emph{\bibinfo{title}{telingo = {ASP} + Time}}.
\newblock In \bibinfo{editor}{Balduccini} et~al.  \cite{lpnmr19}, pp.
  \bibinfo{pages}{256--269}, \doi{10.1007/978-3-030-20528-7\_19}.

\bibitemdeclare{inproceedings}{cakaossc16a}
\bibitem{cakaossc16a}
\bibinfo{author}{P.~\surnamestart Cabalar\surnameend},
  \bibinfo{author}{R.~\surnamestart Kaminski\surnameend},
  \bibinfo{author}{M.~\surnamestart Ostrowski\surnameend} \&
  \bibinfo{author}{T.~\surnamestart Schaub\surnameend} (\bibinfo{year}{2016}):
  \emph{\bibinfo{title}{An {ASP} Semantics for Default Reasoning with
  Constraints}}.
\newblock In \bibinfo{editor}{R.~\surnamestart Kambhampati\surnameend}, editor:
  {\sl \bibinfo{booktitle}{Proceedings of the Twenty-fifth International Joint
  Conference on Artificial Intelligence (IJCAI'16)}},
  \bibinfo{publisher}{IJCAI/AAAI Press}, pp. \bibinfo{pages}{1015--1021},
  \doi{10.5555/3060621.3060762}.

\bibitemdeclare{article}{chkost81a}
\bibitem{chkost81a}
\bibinfo{author}{A.~\surnamestart Chandra\surnameend},
  \bibinfo{author}{D.~\surnamestart Kozen\surnameend} \&
  \bibinfo{author}{L.~\surnamestart Stockmeyer\surnameend}
  (\bibinfo{year}{1981}): \emph{\bibinfo{title}{Alternation}}.
\newblock {\sl \bibinfo{journal}{Journal of the ACM}}
  \bibinfo{volume}{28}(\bibinfo{number}{1}), pp. \bibinfo{pages}{114--133},
  \doi{10.1145/322234.322243}.

\bibitemdeclare{inproceedings}{giavar13a}
\bibitem{giavar13a}
\bibinfo{author}{G.~\surnamestart {De Giacomo}\surnameend} \&
  \bibinfo{author}{M.~\surnamestart Vardi\surnameend} (\bibinfo{year}{2013}):
  \emph{\bibinfo{title}{Linear Temporal Logic and Linear Dynamic Logic on
  Finite Traces}}.
\newblock In \bibinfo{editor}{F.~\surnamestart Rossi\surnameend}, editor: {\sl
  \bibinfo{booktitle}{Proceedings of the Twenty-third International Joint
  Conference on Artificial Intelligence (IJCAI'13)}},
  \bibinfo{publisher}{IJCAI/AAAI Press}, pp. \bibinfo{pages}{854--860}.
\newblock \urlprefix\url{https://www.ijcai.org/Abstract/13/132}.

\bibitemdeclare{inproceedings}{giavar15a}
\bibitem{giavar15a}
\bibinfo{author}{G.~\surnamestart {De Giacomo}\surnameend} \&
  \bibinfo{author}{M.~\surnamestart Vardi\surnameend} (\bibinfo{year}{2015}):
  \emph{\bibinfo{title}{Synthesis for {LTL} and {LDL} on Finite Traces}}.
\newblock In \bibinfo{editor}{Q.~\surnamestart Yang\surnameend} \&
  \bibinfo{editor}{M.~\surnamestart Wooldridge\surnameend}, editors: {\sl
  \bibinfo{booktitle}{Proceedings of the Twenty-fourth International Joint
  Conference on Artificial Intelligence (IJCAI'15)}},
  \bibinfo{publisher}{{AAAI} Press}, pp. \bibinfo{pages}{1558--1564}.
\newblock \urlprefix\url{https://ijcai.org/Abstract/15/223}.

\bibitemdeclare{book}{degola16a}
\bibitem{degola16a}
\bibinfo{author}{S.~\surnamestart Demri\surnameend},
  \bibinfo{author}{V.~\surnamestart Goranko\surnameend} \&
  \bibinfo{author}{M.~\surnamestart Lange\surnameend} (\bibinfo{year}{2016}):
  \emph{\bibinfo{title}{Temporal Logics in Computer Science: Finite-State
  Systems}}.
\newblock \bibinfo{series}{Cambridge Tracts in Theoretical Computer Science},
  \bibinfo{publisher}{Cambridge University Press},
  \doi{10.1017/CBO9781139236119}.

\bibitemdeclare{inproceedings}{fukotamo86a}
\bibitem{fukotamo86a}
\bibinfo{author}{M.~\surnamestart Fujita\surnameend},
  \bibinfo{author}{S.~\surnamestart Kono\surnameend},
  \bibinfo{author}{H.~\surnamestart Tanaka\surnameend} \&
  \bibinfo{author}{T.~\surnamestart Moto{-}Oka\surnameend}
  (\bibinfo{year}{1986}): \emph{\bibinfo{title}{Tokio: Logic Programming
  Language Based on Temporal Logic and its Compilation to {P}rolog}}.
\newblock In \bibinfo{editor}{E.~\surnamestart Shapiro\surnameend}, editor:
  {\sl \bibinfo{booktitle}{Proceedings of the Third International Conference on
  Logic Programming (ICLP'86)}}, {\sl \bibinfo{series}{Lecture Notes in
  Computer Science}} \bibinfo{volume}{225}, \bibinfo{publisher}{Springer}, pp.
  \bibinfo{pages}{695--709}, \doi{10.1007/3-540-16492-8\_119}.

\bibitemdeclare{incollection}{gabbay87b}
\bibitem{gabbay87b}
\bibinfo{author}{D.~\surnamestart Gabbay\surnameend} (\bibinfo{year}{1987}):
  \emph{\bibinfo{title}{Modal and Temporal Logic Programming}}.
\newblock In \bibinfo{editor}{A.~\surnamestart Galton\surnameend}, editor: {\sl
  \bibinfo{booktitle}{Temporal Logics and their Applications}},
  chapter~\bibinfo{chapter}{6}, \bibinfo{publisher}{Academic Press}, pp.
  \bibinfo{pages}{197--237}.

\bibitemdeclare{article}{gekakasc17a}
\bibitem{gekakasc17a}
\bibinfo{author}{M.~\surnamestart Gebser\surnameend},
  \bibinfo{author}{R.~\surnamestart Kaminski\surnameend},
  \bibinfo{author}{B.~\surnamestart Kaufmann\surnameend} \&
  \bibinfo{author}{T.~\surnamestart Schaub\surnameend} (\bibinfo{year}{2019}):
  \emph{\bibinfo{title}{Multi-shot {ASP} solving with clingo}}.
\newblock {\sl \bibinfo{journal}{Theory and Practice of Logic Programming}}
  \bibinfo{volume}{19}(\bibinfo{number}{1}), pp. \bibinfo{pages}{27--82},
  \doi{10.1017/S1471068418000054}.

\bibitemdeclare{inproceedings}{giarub18a}
\bibitem{giarub18a}
\bibinfo{author}{G.~De \surnamestart Giacomo\surnameend} \&
  \bibinfo{author}{S.~\surnamestart Rubin\surnameend} (\bibinfo{year}{2018}):
  \emph{\bibinfo{title}{Automata-Theoretic Foundations of FOND Planning for
  {LTLf} and {LDLf} Goals}}.
\newblock In \bibinfo{editor}{J.~\surnamestart Lang\surnameend}, editor: {\sl
  \bibinfo{booktitle}{Proceedings of the Twenty-seventh International Joint
  Conference on Artificial Intelligence (IJCAI'18)}},
  \bibinfo{publisher}{ijcai.org}, pp. \bibinfo{pages}{4729--4735},
  \doi{10.24963/ijcai.2018/657}.

\bibitemdeclare{article}{hasascst24a}
\bibitem{hasascst24a}
\bibinfo{author}{S.~\surnamestart Hahn\surnameend},
  \bibinfo{author}{O.~\surnamestart Sabuncu\surnameend},
  \bibinfo{author}{T.~\surnamestart Schaub\surnameend} \&
  \bibinfo{author}{T.~\surnamestart Stolzmann\surnameend}
  (\bibinfo{year}{2024}): \emph{\bibinfo{title}{Clingraph: A System for
  ASP-based Visualization}}.
\newblock {\sl \bibinfo{journal}{Theory and Practice of Logic Programming}},
  \doi{10.1017/S147106842400005X}.

\bibitemdeclare{book}{hatiko00a}
\bibitem{hatiko00a}
\bibinfo{author}{D.~\surnamestart Harel\surnameend},
  \bibinfo{author}{J.~\surnamestart Tiuryn\surnameend} \&
  \bibinfo{author}{D.~\surnamestart Kozen\surnameend} (\bibinfo{year}{2000}):
  \emph{\bibinfo{title}{Dynamic Logic}}.
\newblock \bibinfo{publisher}{MIT Press}, \doi{10.1145/568438.568456}.

\bibitemdeclare{inproceedings}{hejejoklparasa95a}
\bibitem{hejejoklparasa95a}
\bibinfo{author}{J.~\surnamestart Henriksen\surnameend},
  \bibinfo{author}{J.~\surnamestart Jensen\surnameend},
  \bibinfo{author}{M.~\surnamestart J{\o}rgensen\surnameend},
  \bibinfo{author}{N.~\surnamestart Klarlund\surnameend},
  \bibinfo{author}{R.~\surnamestart Paige\surnameend},
  \bibinfo{author}{T.~\surnamestart Rauhe\surnameend} \&
  \bibinfo{author}{A.~\surnamestart Sandholm\surnameend}
  (\bibinfo{year}{1995}): \emph{\bibinfo{title}{Mona: Monadic Second-Order
  Logic in Practice}}.
\newblock In \bibinfo{editor}{E.~\surnamestart Brinksma\surnameend},
  \bibinfo{editor}{R.~\surnamestart Cleaveland\surnameend},
  \bibinfo{editor}{K.~\surnamestart Larsen\surnameend},
  \bibinfo{editor}{T.~\surnamestart Margaria\surnameend} \&
  \bibinfo{editor}{B.~\surnamestart Steffen\surnameend}, editors: {\sl
  \bibinfo{booktitle}{Proceedings of the First International Workshop on Tools
  and Algorithms for Construction and Analysis of Systems (TACAS'95)}}, {\sl
  \bibinfo{series}{Lecture Notes in Computer Science}} \bibinfo{volume}{1019},
  \bibinfo{publisher}{Springer-Verlag}, pp. \bibinfo{pages}{89--110},
  \doi{10.1007/3-540-60630-0\_5}.

\bibitemdeclare{incollection}{heyting30a}
\bibitem{heyting30a}
\bibinfo{author}{A.~\surnamestart Heyting\surnameend} (\bibinfo{year}{1930}):
  \emph{\bibinfo{title}{Die formalen {R}egeln der intuitionistischen {L}ogik}}.
\newblock In: {\sl \bibinfo{booktitle}{Sitzungsberichte der Preussischen
  Akademie der Wissenschaften}}, \bibinfo{publisher}{Deutsche Akademie der
  Wissenschaften zu Berlin}, pp. \bibinfo{pages}{42--56}.

\bibitemdeclare{book}{hopull79a}
\bibitem{hopull79a}
\bibinfo{author}{J.~\surnamestart Hopcroft\surnameend} \&
  \bibinfo{author}{J~\surnamestart Ullman\surnameend} (\bibinfo{year}{1979}):
  \emph{\bibinfo{title}{Introduction to Automata Theory, Languages, and
  Computation}}.
\newblock \bibinfo{publisher}{Addison-Wesley}.

\bibitemdeclare{article}{huozdi20a}
\bibitem{huozdi20a}
\bibinfo{author}{U.~\surnamestart Hustadt\surnameend},
  \bibinfo{author}{A.~\surnamestart Ozaki\surnameend} \&
  \bibinfo{author}{C.~\surnamestart Dixon\surnameend} (\bibinfo{year}{2020}):
  \emph{\bibinfo{title}{Theorem Proving for Pointwise Metric Temporal Logic
  Over the Naturals via Translations}}.
\newblock {\sl \bibinfo{journal}{Journal of Automated Reasoning}}
  \bibinfo{volume}{64}(\bibinfo{number}{8}), pp. \bibinfo{pages}{1553--1610},
  \doi{10.1007/s10817-020-09541-4}.

\bibitemdeclare{article}{jakaosscscwa17a}
\bibitem{jakaosscscwa17a}
\bibinfo{author}{T.~\surnamestart Janhunen\surnameend},
  \bibinfo{author}{R.~\surnamestart Kaminski\surnameend},
  \bibinfo{author}{M.~\surnamestart Ostrowski\surnameend},
  \bibinfo{author}{T.~\surnamestart Schaub\surnameend},
  \bibinfo{author}{S.~\surnamestart Schellhorn\surnameend} \&
  \bibinfo{author}{P.~\surnamestart Wanko\surnameend} (\bibinfo{year}{2017}):
  \emph{\bibinfo{title}{Clingo goes Linear Constraints over Reals and
  Integers}}.
\newblock {\sl \bibinfo{journal}{Theory and Practice of Logic Programming}}
  \bibinfo{volume}{17}(\bibinfo{number}{5-6}), pp. \bibinfo{pages}{872--888},
  \doi{10.1017/S1471068417000242}.

\bibitemdeclare{article}{jiakum04a}
\bibitem{jiakum04a}
\bibinfo{author}{S.~\surnamestart Jiang\surnameend} \&
  \bibinfo{author}{R.~\surnamestart Kumar\surnameend} (\bibinfo{year}{2004}):
  \emph{\bibinfo{title}{Failure diagnosis of discrete-event systems with
  linear-time temporal logic specifications}}.
\newblock {\sl \bibinfo{journal}{IEEE Transactions on Automatic Control}}
  \bibinfo{volume}{49}(\bibinfo{number}{6}), pp. \bibinfo{pages}{934--945},
  \doi{10.1109/TAC.2004.829616}.

\bibitemdeclare{inproceedings}{kascwa17a}
\bibitem{kascwa17a}
\bibinfo{author}{R.~\surnamestart Kaminski\surnameend},
  \bibinfo{author}{T.~\surnamestart Schaub\surnameend} \&
  \bibinfo{author}{P.~\surnamestart Wanko\surnameend} (\bibinfo{year}{2017}):
  \emph{\bibinfo{title}{A Tutorial on Hybrid Answer Set Solving with clingo}}.
\newblock In \bibinfo{editor}{G.~\surnamestart Ianni\surnameend},
  \bibinfo{editor}{D.~\surnamestart Lembo\surnameend},
  \bibinfo{editor}{L.~\surnamestart Bertossi\surnameend},
  \bibinfo{editor}{W.~\surnamestart Faber\surnameend},
  \bibinfo{editor}{B.~\surnamestart Glimm\surnameend},
  \bibinfo{editor}{G.~\surnamestart Gottlob\surnameend} \&
  \bibinfo{editor}{S.~\surnamestart Staab\surnameend}, editors: {\sl
  \bibinfo{booktitle}{Proceedings of the Thirteenth International Summer School
  of the Reasoning Web}}, {\sl \bibinfo{series}{Lecture Notes in Computer
  Science}} \bibinfo{volume}{10370}, \bibinfo{publisher}{Springer-Verlag}, pp.
  \bibinfo{pages}{167--203}, \doi{10.1007/978-3-319-61033-7\_6}.

\bibitemdeclare{phdthesis}{kamp68a}
\bibitem{kamp68a}
\bibinfo{author}{J.~\surnamestart Kamp\surnameend} (\bibinfo{year}{1968}):
  \emph{\bibinfo{title}{Tense Logic and the Theory of Linear Order}}.
\newblock Ph.D. thesis, \bibinfo{school}{University of California at Los
  Angeles}.

\bibitemdeclare{article}{kapzak20a}
\bibitem{kapzak20a}
\bibinfo{author}{C.~\surnamestart Kapoutsis\surnameend} \&
  \bibinfo{author}{M.~\surnamestart Zakzok\surnameend} (\bibinfo{year}{2021}):
  \emph{\bibinfo{title}{Alternation in two-way finite automata}}.
\newblock {\sl \bibinfo{journal}{Theoretical Computer Science}}
  \bibinfo{volume}{870}, pp. \bibinfo{pages}{75--102},
  \doi{10.1016/j.tcs.2020.12.011}.

\bibitemdeclare{inproceedings}{karfra08a}
\bibitem{karfra08a}
\bibinfo{author}{S.~\surnamestart Karaman\surnameend} \&
  \bibinfo{author}{E.~\surnamestart Frazzoli\surnameend}
  (\bibinfo{year}{2008}): \emph{\bibinfo{title}{Vehicle routing problem with
  metric temporal logic specifications}}.
\newblock In: {\sl \bibinfo{booktitle}{2008 47th IEEE conference on decision
  and control}}, \bibinfo{organization}{IEEE}, pp. \bibinfo{pages}{3953--3958},
  \doi{10.1109/CDC.2008.4739366}.

\bibitemdeclare{article}{koymans90a}
\bibitem{koymans90a}
\bibinfo{author}{R.~\surnamestart Koymans\surnameend} (\bibinfo{year}{1990}):
  \emph{\bibinfo{title}{Specifying Real-Time Properties with Metric Temporal
  Logic}}.
\newblock {\sl \bibinfo{journal}{Real-Time Systems}}
  \bibinfo{volume}{2}(\bibinfo{number}{4}), pp. \bibinfo{pages}{255--299},
  \doi{10.1007/BF01995674}.

\bibitemdeclare{article}{laliso84a}
\bibitem{laliso84a}
\bibinfo{author}{R.~\surnamestart Ladner\surnameend},
  \bibinfo{author}{R.~\surnamestart Lipton\surnameend} \&
  \bibinfo{author}{L~\surnamestart Stockmeyer\surnameend}
  (\bibinfo{year}{1984}): \emph{\bibinfo{title}{Alternating pushdown and stack
  automata}}.
\newblock {\sl \bibinfo{journal}{SIAM Journal on Computing}}
  \bibinfo{volume}{13}(\bibinfo{number}{1}), pp. \bibinfo{pages}{135--155},
  \doi{10.1137/0213010}.

\bibitemdeclare{article}{lereleli97a}
\bibitem{lereleli97a}
\bibinfo{author}{H.~\surnamestart Levesque\surnameend},
  \bibinfo{author}{R.~\surnamestart Reiter\surnameend},
  \bibinfo{author}{Y.~\surnamestart Lesp{\'e}rance\surnameend},
  \bibinfo{author}{F.~\surnamestart Lin\surnameend} \&
  \bibinfo{author}{R.~\surnamestart Scherl\surnameend} (\bibinfo{year}{1997}):
  \emph{\bibinfo{title}{{GOLOG}: A Logic Programming Language for Dynamic
  Domains.}}
\newblock {\sl \bibinfo{journal}{Journal of Logic Programming}}
  \bibinfo{volume}{31}(\bibinfo{number}{1-3}), pp. \bibinfo{pages}{59--83},
  \doi{10.1016/S0743-1066(96)00121-5}.

\bibitemdeclare{inproceedings}{luvalibemc16a}
\bibitem{luvalibemc16a}
\bibinfo{author}{R.~\surnamestart Luo\surnameend},
  \bibinfo{author}{R.~\surnamestart Valenzano\surnameend},
  \bibinfo{author}{Y.~\surnamestart Li\surnameend},
  \bibinfo{author}{C.~\surnamestart Beck\surnameend} \&
  \bibinfo{author}{S.~\surnamestart McIlraith\surnameend}
  (\bibinfo{year}{2016}): \emph{\bibinfo{title}{Using Metric Temporal Logic to
  Specify Scheduling Problems}}.
\newblock In \bibinfo{editor}{C.~\surnamestart Baral\surnameend},
  \bibinfo{editor}{J.~\surnamestart Delgrande\surnameend} \&
  \bibinfo{editor}{F.~\surnamestart Wolter\surnameend}, editors: {\sl
  \bibinfo{booktitle}{Proceedings of the Fifteenth International Conference on
  Principles of Knowledge Representation and Reasoning (KR'16)}},
  \bibinfo{publisher}{{AAAI} Press}, pp. \bibinfo{pages}{581--584}.
\newblock \urlprefix\url{https://www.aaai.org/ocs/index.php/KR/KR16/paper/view/12909}.

\bibitemdeclare{inproceedings}{nivpit10a}
\bibitem{nivpit10a}
\bibinfo{author}{D.~\surnamestart Ni\v{c}kovi\'c\surnameend} \&
  \bibinfo{author}{N.~\surnamestart Piterman\surnameend}
  (\bibinfo{year}{2010}): \emph{\bibinfo{title}{From {MTL} to Deterministic
  Timed Automata}}.
\newblock In \bibinfo{editor}{K.~\surnamestart Chatterjee\surnameend} \&
  \bibinfo{editor}{T.~\surnamestart Henzinger\surnameend}, editors: {\sl
  \bibinfo{booktitle}{Proceedings of the Eighth International Conference on
  Formal Modeling and Analysis of Timed Systems (FORMATS'10)}},
  \bibinfo{series}{Lecture Notes in Computer Science},
  \bibinfo{publisher}{Springer-Verlag}, pp. \bibinfo{pages}{152--167},
  \doi{10.1007/978-3-642-15297-9\_13}.

\bibitemdeclare{inproceedings}{orgwad92a}
\bibitem{orgwad92a}
\bibinfo{author}{M.~\surnamestart Orgun\surnameend} \&
  \bibinfo{author}{W.~\surnamestart Wadge\surnameend} (\bibinfo{year}{1992}):
  \emph{\bibinfo{title}{Theory and Practice of Temporal Logic Programming}}.
\newblock In \bibinfo{editor}{L.~\surnamestart Fariñas del Cerro\surnameend} \&
\bibinfo{editor}{M.~\surnamestart Penttonen\surnameend}, editors: {\sl
\bibinfo{booktitle}{Intensional Logics for Programming}},
chapter~\bibinfo{chapter}{2},
\bibinfo{publisher}{Oxford University Press}, pp. \bibinfo{pages}{21--50},
\doi{10.1093/oso/9780198537755.003.0002}.

\bibitemdeclare{article}{pearce06a}
\bibitem{pearce06a}
\bibinfo{author}{D.~\surnamestart Pearce\surnameend} (\bibinfo{year}{2006}):
  \emph{\bibinfo{title}{Equilibrium logic}}.
\newblock {\sl \bibinfo{journal}{Annals of Mathematics and Artificial
  Intelligence}} \bibinfo{volume}{47}(\bibinfo{number}{1-2}), pp.
  \bibinfo{pages}{3--41}, \doi{10.1007/s10472-006-9028-z}.

\bibitemdeclare{article}{smivar21a}
\bibitem{smivar21a}
\bibinfo{author}{K.~\surnamestart Smith\surnameend} \&
  \bibinfo{author}{M.~\surnamestart Vardi\surnameend} (\bibinfo{year}{2021}):
  \emph{\bibinfo{title}{Automata Linear Dynamic Logic on Finite Traces}}.
\newblock {\sl \bibinfo{journal}{arXiv preprint arXiv:2108.12003}}.

\bibitemdeclare{article}{tseitin68a}
\bibitem{tseitin68a}
\bibinfo{author}{G.~\surnamestart Tseitin\surnameend} (\bibinfo{year}{1968}):
  \emph{\bibinfo{title}{On the complexity of derivation in the propositional
  calculus}}.
\newblock {\sl \bibinfo{journal}{Zapiski nauchnykh seminarov LOMI}}
  \bibinfo{volume}{8}, pp. \bibinfo{pages}{234--259}.

\bibitemdeclare{inproceedings}{vardi95a}
\bibitem{vardi95a}
\bibinfo{author}{M.~\surnamestart Vardi\surnameend} (\bibinfo{year}{1995}):
  \emph{\bibinfo{title}{An Automata-Theoretic Approach to Linear Temporal
  Logic}}.
\newblock In \bibinfo{editor}{F.~\surnamestart Moller\surnameend} \&
  \bibinfo{editor}{G.~\surnamestart Birtwistle\surnameend}, editors: {\sl
  \bibinfo{booktitle}{Logics for Concurrency: Structure versus Automata}}, {\sl
  \bibinfo{series}{Lecture Notes in Computer Science}} \bibinfo{volume}{1043},
  \bibinfo{publisher}{Springer-Verlag}, pp. \bibinfo{pages}{238--266},
  \doi{10.1007/3-540-60915-6\_6}.

\bibitemdeclare{inproceedings}{vardi97a}
\bibitem{vardi97a}
\bibinfo{author}{M.~\surnamestart Vardi\surnameend} (\bibinfo{year}{1997}):
  \emph{\bibinfo{title}{Alternating Automata: Unifying Truth and Validity
  Checking for Temporal Logics}}.
\newblock In \bibinfo{editor}{W.~\surnamestart McCune\surnameend}, editor: {\sl
  \bibinfo{booktitle}{Proceedings of the Fourteenth International Conference on
  Automated Deduction (CADE'97)}}, {\sl \bibinfo{series}{Lecture Notes in
  Computer Science}} \bibinfo{volume}{1249},
  \bibinfo{publisher}{Springer-Verlag}, pp. \bibinfo{pages}{191--206},
  \doi{10.1007/3-540-63104-6\_19}.

\bibitemdeclare{inproceedings}{zhpuva19a}
\bibitem{zhpuva19a}
\bibinfo{author}{S.~\surnamestart Zhu\surnameend},
  \bibinfo{author}{G.~\surnamestart Pu\surnameend} \&
  \bibinfo{author}{M.~\surnamestart Vardi\surnameend} (\bibinfo{year}{2019}):
  \emph{\bibinfo{title}{First-Order vs. Second-Order Encodings for
  {LTLf}-to-Automata Translation}}.
\newblock In \bibinfo{editor}{T.~\surnamestart Gopal\surnameend} \&
  \bibinfo{editor}{J.~\surnamestart Watada\surnameend}, editors: {\sl
  \bibinfo{booktitle}{Proceedings of the Fifteenth Annual Conference on Theory
  and Applications of Models of Computation (TAMC'19)}}, {\sl
  \bibinfo{series}{Lecture Notes in Computer Science}} \bibinfo{volume}{11436},
  \bibinfo{publisher}{Springer-Verlag}, pp. \bibinfo{pages}{684--705}, \doi{10.1007/978-3-030-14812-6\_43}.

\end{thebibliography}

\end{document}